\documentclass{article}

     \usepackage[preprint]{neurips_2022}

\usepackage[utf8]{inputenc} 
\usepackage[T1]{fontenc}    
\usepackage{hyperref}       
\usepackage{url}            
\usepackage{booktabs}       
\usepackage{amsfonts}       
\usepackage{nicefrac}       
\usepackage{microtype}      
\usepackage{xcolor}         
\usepackage{graphicx}
\usepackage{wrapfig}
\usepackage{amsmath}
\usepackage{amssymb}
\usepackage{mathtools}
\usepackage{booktabs}
\usepackage{multirow}
\usepackage{threeparttable}
\usepackage{verbatim}
\usepackage{multicol}
\usepackage{bm}
\usepackage{algorithm}
\usepackage{algorithmic}
\usepackage{subfigure}
\usepackage{diagbox}

\usepackage{longtable}
\usepackage{caption}

\title{Survey on Monocular Metric Depth Estimation}

\author{%
  Jiuling Zhang\\
  	University of Chinese Academy of Sciences\\
  \texttt{zhangjiuling19@mails.ucas.edu.cn} \\
\\
}

\begin{document}

\maketitle

\begin{abstract}
Monocular Depth Estimation (MDE) enables spatial understanding, 3D reconstruction, and autonomous navigation, yet deep learning approaches often predict only relative depth without a consistent metric scale. This limitation reduces reliability in applications such as visual SLAM, precise 3D modeling, and view synthesis. Monocular Metric Depth Estimation (MMDE) overcomes this challenge by producing depth maps with absolute scale, ensuring geometric consistency and enabling deployment without additional calibration. This survey reviews the evolution of MMDE, from geometry-based methods to state-of-the-art deep models, with emphasis on the datasets that drive progress. Key benchmarks, including KITTI, NYU-D, ApolloScape, and TartanAir, are examined in terms of modality, scene type, and application domain. Methodological advances are analyzed, covering domain generalization, boundary preservation, and the integration of synthetic and real data. Techniques such as unsupervised and semi-supervised learning, patch-based inference, architectural innovations, and generative modeling are evaluated for their strengths and limitations. By synthesizing current progress, highlighting the importance of high-quality datasets, and identifying open challenges, this survey provides a structured reference for advancing MMDE and supporting its adoption in real-world computer vision systems.
\end{abstract}

\section{Preliminary}
Depth estimation reconstructs 3D scene structure from images and underpins numerous applications, including 3D reconstruction~\citep{mildenhall2021nerf,kerbl20233d,ye2024gaustudio}, autonomous navigation~\citep{szeliski2022computer}, self-driving vehicles~\citep{zheng2024physical}, and video understanding~\citep{leduc2024soccernet}. It also supports emerging areas such as AI-generated content (AIGC), which spans image synthesis~\citep{zhang2023adding,khan2023tiled}, video generation~\citep{liew2023magicedit}, and 3D scene reconstruction~\citep{xu2023neurallift,shahbazi2024inserf,shriram2024realmdreamer}.

Classical approaches relied on parallax, stereo vision, and multi-camera systems. With advances in deep learning, monocular depth estimation (MDE) emerged as a cost-effective alternative that predicts depth from a single image. The field’s growing significance is reflected in the Monocular Depth Estimation Challenge (MDEC), hosted at CVPR in 2023 and 2024 and scheduled to return in 2025\footnote{https://jspenmar.github.io/MDEC/}.

Recent work has shifted toward Monocular Metric Depth Estimation (MMDE), which provides absolute depth values rather than scale-inconsistent maps. This shift is driven by practical requirements for accurate, generalizable, and detail-preserving predictions in real-world tasks. Industry leaders, including Intel~\citep{bhat2023zoedepth}, Apple~\citep{bochkovskii2024depth}, DeepMind~\citep{saxena2023zero}, TikTok~\citep{yang2024depth1,yang2024depth2}, and Bosch~\citep{guo2025depth}, have advanced MMDE through large-scale datasets, high-performance computing, and novel architectures, enabling improvements in zero-shot generalization and reconstruction fidelity.

However, survey literature remains limited. Most comprehensive reviews predate 2020~\citep{bhoi2019monocular,khan2020deep,zhao2020monocular,xiaogang2020monocular}, and recent work often focuses on domain-specific settings~\citep{lahiri2024deep,tosi2024diffusion,vyas2022outdoor,dong2022towards} or relative depth~\citep{masoumian2022monocular,arampatzakis2023monocular,rajapaksha2024deep}. Meanwhile, leading venues such as CVPR 2024, ECCV 2024, and NeurIPS 2024 highlight emerging trends in zero-shot MMDE and generative model integration. This paper fills a critical gap by offering a comprehensive review of MMDE, addressing datasets, methodological advances, open challenges, and future directions.

\section{Depth Estimation}
The goal of depth estimation is to compute a depth map $D := \left(\mathbb{R}\right)^{H \times W}$ from a given 2D image $\mathit{I} := \left(\mathbb{R}\right)^{H \times W \times 3}$, where each depth value $d_{i , j} \in D$ represents the physical distance between pixel $i_{i , j} \in I$ and the camera~\citep{bhat2021adabins}. This task is inherently ill-posed because 2D images are projections of the 3D world, which discard geometric information. Monocular depth estimation is particularly challenging due to the absence of parallax and auxiliary cues~\citep{miangoleh2021boosting}.

Despite these challenges, depth estimation plays a critical role in computer vision~\citep{jampani2021slide}. Accurate depth maps enhance scene understanding and object localization, which benefit a wide range of applications. In autonomous driving and robotics, they improve obstacle detection, path planning, and environmental awareness. In AR/VR, reliable depth predictions enable realistic 3D reconstruction and immersive interaction. In computational photography, depth supports multi-focus imaging, background segmentation, and 3D video synthesis~\citep{eigen2014depth}.

By providing dense, pixel-wise distance predictions, depth estimation equips intelligent systems with geometric awareness of the environment. This capability remains a cornerstone of visual perception research with substantial real-world impact across established and emerging domains.

\subsection{Traditional Methods}

Before the rise of deep learning, depth estimation primarily relied on geometric models and specialized sensors. These approaches achieved accuracy in controlled conditions but often required additional hardware and struggled in complex real-world environments~\citep{singh2023depth}.

\subsubsection{Sensors}
Early sensor-based systems directly captured spatial information. Structured-light devices, such as Microsoft Kinect v1, projected predefined patterns to infer depth, while Time-of-Flight (ToF) sensors measured light travel time. Although accurate in laboratory settings, these methods were expensive and highly sensitive to ambient light and surface properties, limiting their use in dynamic or portable applications.

\subsubsection{Stereo Vision}
Stereo vision, inspired by human binocular perception, estimated depth from disparities between two calibrated cameras. While effective, performance degraded in low-texture regions, poor lighting, and dynamic scenes, where reliable pixel correspondence was difficult. The hardware complexity of stereo rigs further restricted widespread deployment.

\subsubsection{Geometric Multi-Frame Methods}
Techniques such as Structure-from-Motion (SfM) and Simultaneous Localization and Mapping (SLAM) inferred depth from multi-frame parallax by estimating camera poses and reconstructing 3D point clouds. Indirect methods minimized reprojection error from feature correspondences, while direct methods exploited photometric consistency~\citep{wofk2023monocular}. Despite enabling depth estimation without extra sensors, these methods were sensitive to illumination changes and texture inconsistencies, reducing robustness in unconstrained environments.

Traditional approaches provided a strong foundation but were limited by hardware demands, environmental constraints, and computational complexity. The shift to deep learning introduced more scalable, flexible, and robust solutions, which now dominate depth estimation research.

\subsection{Deep Learning}

Deep learning has fundamentally reshaped depth estimation, replacing geometry-based methods with learning-driven approaches. Unlike stereo or LiDAR-based systems, neural networks predict depth directly from a single image, reducing hardware cost and enabling lightweight deployment in applications such as mobile AR and drone navigation~\citep{garg2016unsupervised}.

A major advantage of deep learning is its ability to leverage large-scale datasets to capture scene priors that resolve the inherent ambiguity of monocular input. Neural networks learn both local textures and global semantics, allowing inference of spatial relationships and geometric structure. For example, sky regions are recognized as distant, while ground-plane textures provide depth gradients, leading to more reliable predictions even in ill-posed regions.

Feature representation is central to modern models. Convolution Neural Networks (CNNs) extract multi-scale information, integrating fine textures with high-level semantics to improve precision and robustness in structured environments. This feature-driven paradigm surpasses pixel-based geometric methods by achieving reliable performance across sparse-texture, cluttered, and dynamic scenes.

Deep learning thus enables scalable, cost-efficient, and robust monocular depth estimation, expanding its impact on domains such as autonomous driving, robotics, and immersive AR/VR systems.

\section{Monocular Depth Estimation}

Monocular Depth Estimation (MDE) predicts scene depth from a single RGB image, eliminating the need for multi-view setups or specialized sensors. Neural networks extract visual cues directly, thereby reducing system complexity and cost compared to traditional geometry-based methods.

Early supervised approaches demonstrated the feasibility of learning depth from labeled datasets. Eigen et al. introduced a multi-scale CNN that jointly predicted global and local depth, significantly improving accuracy~\citep{eigen2014depth}. Their subsequent work incorporated surface normals and semantic labels in a multi-task framework, further enhancing robustness~\citep{eigen2015predicting}.

Modern architectures largely adopt encoder–decoder designs, where encoders capture global context and decoders reconstruct fine-grained depth. Multi-scale feature fusion strengthens the balance between structure and detail. To mitigate the intrinsic ambiguity of monocular cues, several works integrate geometric priors such as perspective constraints and object size, improving plausibility and generalization.

Although early models often struggled with cross-domain transfer, advances in universal feature extraction and domain-invariant learning have expanded MDE’s applicability in real-world scenarios. These developments establish MDE as a practical and scalable solution for tasks requiring dense depth perception.

\section{Zero-shot Monocular Depth Estimation}

Zero-shot MDE addresses the poor transferability of supervised models, which often rely on dataset-specific scales and camera intrinsics. Early solutions reformulated the task as Relative Depth Estimation (RDE), predicting ordinal pixel relationships rather than absolute distances. This scale-agnostic formulation, coupled with scale-invariant and scale-and-shift-invariant loss functions, improved generalization across heterogeneous datasets~\citep{fu2018deep}.

A major milestone was MiDaS~\citep{birkl2023midas}, which unified multi-dataset training under scale-invariant objectives. By evolving from CNN-based designs to Vision Transformers~\citep{han2022survey}, MiDaS demonstrated strong cross-domain performance, establishing a foundation for zero-shot depth prediction. However, RDE inherently trades metric precision for generalization: while robust across domains, it cannot provide absolute scale, limiting applications such as SLAM, AR, and autonomous driving where metric consistency and temporal coherence are critical.

Recent research aims to bridge relative and metric estimation within unified frameworks, seeking to balance robustness with scale fidelity. These efforts mark a shift toward zero-shot models that can generalize across domains while remaining suitable for real-world deployment.

\section{Monocular Metric Depth Estimation}

Monocular Metric Depth Estimation (MMDE) has gained prominence due to its ability to predict absolute depth in physical units, enabling consistent 3D perception for applications such as reconstruction, novel view synthesis, and SLAM. Unlike relative methods, MMDE ensures geometric stability and temporal coherence across frames, making it more practical for real-world deployment.

Early approaches assumed known camera intrinsics. Metric3D mapped images to a canonical space with focal length corrections~\citep{yin2023metric3d}, while ZeroDepth leveraged variational inference with camera-specific embeddings~\citep{guizilini2023towards}. More recent work removes this dependency by estimating intrinsics through auxiliary networks or predicting depth in spherical representations~\citep{spencer2024kick}.

Advances in adaptive binning have further improved accuracy. AdaBins introduced dynamic depth bin allocation~\citep{bhat2021adabins}, refined by LocalBins through spatial partitioning~\citep{bhat2022localbins}, while BinsFormer integrated Transformers for global-local bin optimization~\citep{li2024binsformer}. NeW CRFs combined neural networks with conditional random fields to enforce pixel-wise consistency~\citep{yuan2022new}.

A breakthrough in zero-shot MMDE was ZoeDepth~\citep{bhat2023zoedepth}, which extended MiDaS with adaptive metric binning and scene-aware routing, achieving strong cross-domain generalization across indoor and outdoor datasets. This unified design has established a benchmark for scalable and robust metric depth estimation.

\section{Challenges and Improvements}
Although MMDE has advanced significantly, generalization to unseen domains remains a primary challenge~\citep{spencer2024third}. Models often suffer from geometric blurring, loss of fine details, and degraded performance in high-resolution or cross-domain scenarios, limiting reliability in real-world applications.

Recent improvements target these issues through architectural refinements, such as multi-scale feature fusion and transformer-based designs, which better preserve structure and context. Enhanced training strategies, including domain-invariant learning and large-scale multi-dataset supervision, improve robustness across diverse environments. At inference, patch-based and adaptive mechanisms mitigate resolution constraints while maintaining geometric consistency. Collectively, these advances have boosted prediction accuracy and stability, yet achieving both high precision and strong generalization in dynamic settings remains an open challenge for MMDE.

\subsection{Generalizability}
Improving the generalization of zero-shot MMDE relies on large-scale data augmentation, robust architectures, and novel training paradigms.

\textbf{Dataset Augmentation:} Depth Anything employs a semi-supervised framework generating 62M self-annotated images, enabling strong cross-domain adaptation through semantic priors and large-scale supervision~\citep{yang2024depth1,marsal2024monoprob,haji2024large,shao2024monodiffusion,wang2024sqldepth}. Depth Any Camera (DAC) further extends depth estimation to fisheye and 360° imagery using ERP, pitch-aware conversion, and field-of-view alignment, achieving robust omnidirectional prediction~\citep{guo2025depth}.

\textbf{Model Improvements:} UniDepth predicts metric 3D point clouds without camera intrinsics via a self-promptable camera module, pseudo-spherical outputs, and geometric invariance loss, enhancing robustness to camera variations and domain shifts~\citep{piccinelli2024unidepth}.

\textbf{Loss and Training Paradigms:} DepthAnything-AC introduces unsupervised consistency regularization and a Spatial Distance Constraint, reducing noise sensitivity and improving fine-structure preservation in degraded conditions~\citep{sun2025depth}.

\subsection{Blurriness}
Depth estimation models often suffer from blurred edges and loss of fine details, particularly around object boundaries, occlusions, and high-frequency textures. This degradation reduces structural accuracy and limits applicability in high-precision tasks.

\textbf{Patch-based methods} improve detail by combining local and global cues. BoostingDepth and PatchFusion use multi-resolution fusion to enhance sharpness, though at high computational cost~\citep{miangoleh2021boosting,li2024patchfusion}. PatchRefiner introduces Detail and Scale Disentangling (DSD) loss and pseudo-labeling to sharpen boundaries while improving efficiency~\citep{li2024patchrefiner}. DepthPro further balances detail and speed through a multi-scale Vision Transformer with patch slicing, though with reduced accuracy in distant regions~\citep{bochkovskii2024depth}.

\textbf{Synthetic datasets} provide pixel-accurate labels for training, reducing boundary artifacts. Depth Anything V2 leverages synthetic supervision with pseudo-labeling and gradient-matching loss to bridge domain gaps, enhancing fine detail and robustness~\citep{yang2024depth2,li2024patchrefiner}.

\textbf{Generative diffusion methods} restore structural fidelity through progressive refinement. Marigold achieves sharper predictions in reflective and transparent regions~\citep{ke2024repurposing}, while GeoWizard introduces scene-aware decoupling and normal map integration for improved 3D geometry~\citep{fu2024geowizard}. DeepMind’s DMD applies logarithmic depth parameterization and FOV conditioning to resolve scale ambiguities and accelerate inference~\citep{saxena2023zero}.

\subsection{Incremental Sensor Assistance}
Lightweight sensor cues have been explored as auxiliary signals to improve monocular metric depth estimation. Lin et al.~\citep{lin2025prompting} propose incorporating sparse LiDAR data as incremental prompts within a depth decoder, guiding pre-trained foundation models toward more accurate and high-resolution metric predictions. This hybrid strategy leverages minimal sensor input while retaining the scalability of purely vision-based approaches.

\begin{table}[t]
\caption{Timeline of key advancements in monocular metric depth estimation (MMDE). While most generative approaches focus on relative depth, Diffusion for Metric Depth (DMD) remains the only reported method producing absolute metric predictions~\citep{saxena2023zero}. However, the absence of public release limits independent validation, leaving the role of generative models in MMDE an open avenue for future exploration.}
\label{table1}
\resizebox{\textwidth}{!}{
\begin{tabular}{lllllll}
\toprule
Method                    & Publication & Category       & Inference & Dataset        & Output   & Source \\
\midrule 
Zoedepth~\citep{bhat2023zoedepth}          & Arxiv       & discriminative & single    & real           & metric   & open         \\
Depth Anything~\citep{yang2024depth1} & CVPR '24    & discriminative & single    & real           & metric   & open         \\
Patch Fusion~\citep{li2024patchfusion}      & CVPR '24    & discriminative & multiple  & real           & metric   & open         \\
Unidepth~\citep{piccinelli2024unidepth}          & CVPR '24    & discriminative & single    & real           & metric   & open         \\
Marigold~\citep{ke2024repurposing}          & CVPR '24    & generative     & multiple  & synthetic      & relative & open         \\
DMD~\citep{saxena2023zero}               & Arxiv       & generative     & multiple  & real           & metric   & close          \\
Depth Anything v2~\citep{yang2024depth2} & NeurIPS '24       & discriminative & single    & real+synthetic & metric   & open         \\
GeoWizard~\citep{fu2024geowizard}         & ECCV '24    & generative     & multiple  & real+synthetic & relative & open         \\
Patch Refiner~\citep{li2024patchrefiner}     & ECCV '24    & discriminative & multiple  & real+synthetic & metric   & open         \\
Depth pro~\citep{bochkovskii2024depth}         & Arxiv       & discriminative & multiple  & real+synthetic & metric   & open  \\
DAC~\citep{guo2025depth}         & Arxiv       & discriminative & single  & real+synthetic & metric   & open  \\
Depth Anything AC~\citep{sun2025depth}         & Arxiv       & discriminative & multiple (patching)  & real+synthetic & metric   & open  \\
\bottomrule
\end{tabular}
}
\end{table}

\begin{table}[]
\caption{Comparative evaluation of monocular metric depth estimation (MMDE) models. ZeroDepth is limited by storage, Metric3D depends on camera parameters, and Depth Anything underperforms in zero-shot generalization. Results across six zero-shot (higher-is-better) and two non-zero-shot (AbsRel, lower-is-better) benchmarks, reported by DepthPro~\citep{bochkovskii2024depth}, reveal substantial performance variation. The absence of standardized datasets, training protocols, and model settings continues to hinder fair cross-model comparison.}
\label{table2}
\resizebox{\textwidth}{!}{
\begin{tabular}{lllllllll}
\toprule
\multirow{2}{*}{\diagbox[width=5.5cm, height=0.8cm]{Method}{Dataset}   }       & Booster↑ & ETH3D↑ & Middlebury↑ & NuScenes↑ & Sintel↑ & Sun-RGBD↑ & NYU v2↓ & KITTI↓ \\
                         & indoor       & outdoor     & outdoor          & outdoor        & outdoor      & indoor        & indoor      & outdoor     \\
\midrule 
DepthAnything~\citep{yang2024depth1}     & 52.3     & 9.3    & 39.3        & 35.4      & 6.9     & 85.0      & 4.3     & 7.6    \\
DepthAnything V2~\citep{yang2024depth2} & 59.5     & 36.3   & 37.2        & 17.7      & 5.9     & 72.4      & 4.4     & 7.4    \\
Metric3D~\citep{yin2023metric3d}        & 4.7      & 34.2   & 13.6        & 64.4      & 17.3    & 16.9      & 8.3     & 5.8    \\
Metric3D v2~\citep{hu2024metric3d}     & 39.4     & 87.7   & 29.9        & 82.6      & 38.3    & 75.6      & 4.5     & 3.9    \\
PatchFusion~\citep{li2024patchfusion}      & 22.6     & 51.8   & 49.9        & 20.4      & 14.0    & 53.6      & -       & -      \\
UniDepth~\citep{piccinelli2024unidepth}          & 27.6     & 25.3   & 31.9        & 83.6      & 16.5    & 95.8      & 5.78    & 4.2    \\
ZeroDepth~\citep{bhat2023zoedepth}       & -        & -      & 46.5        & 64.3      & 12.9    & -         & 8.4     & 10.5   \\
ZoeDepth~\citep{bhat2023zoedepth}        & 21.6     & 34.2   & 53.8        & 28.1      & 7.8     & 85.7      & 7.7     & 5.7    \\
Depth Pro~\citep{bochkovskii2024depth}        & 46.6     & 41.5   & 60.5        & 49.1      & 40.0    & 89.0      & -       & -    \\
\bottomrule
\end{tabular}
}
\end{table}

\subsection{Analysis and Comparison}
Single-inference methods dominate monocular depth estimation due to their efficiency, producing depth in a single forward pass suitable for real-time tasks such as navigation and view synthesis. However, these models often lose fine structural details and rely heavily on large-scale, high-quality annotations, limiting generalization in noisy real-world datasets.

Patch-based strategies improve resolution by processing local regions independently and fusing results, enabling finer structural recovery. Methods like PatchFusion enhance accuracy via optimized patch weighting, but inference scales with patch count, causing latency that restricts deployment in high-resolution or time-critical scenarios~\citep{li2024patchfusion}.

\begin{figure*}
\centerline{\includegraphics[width=35pc]{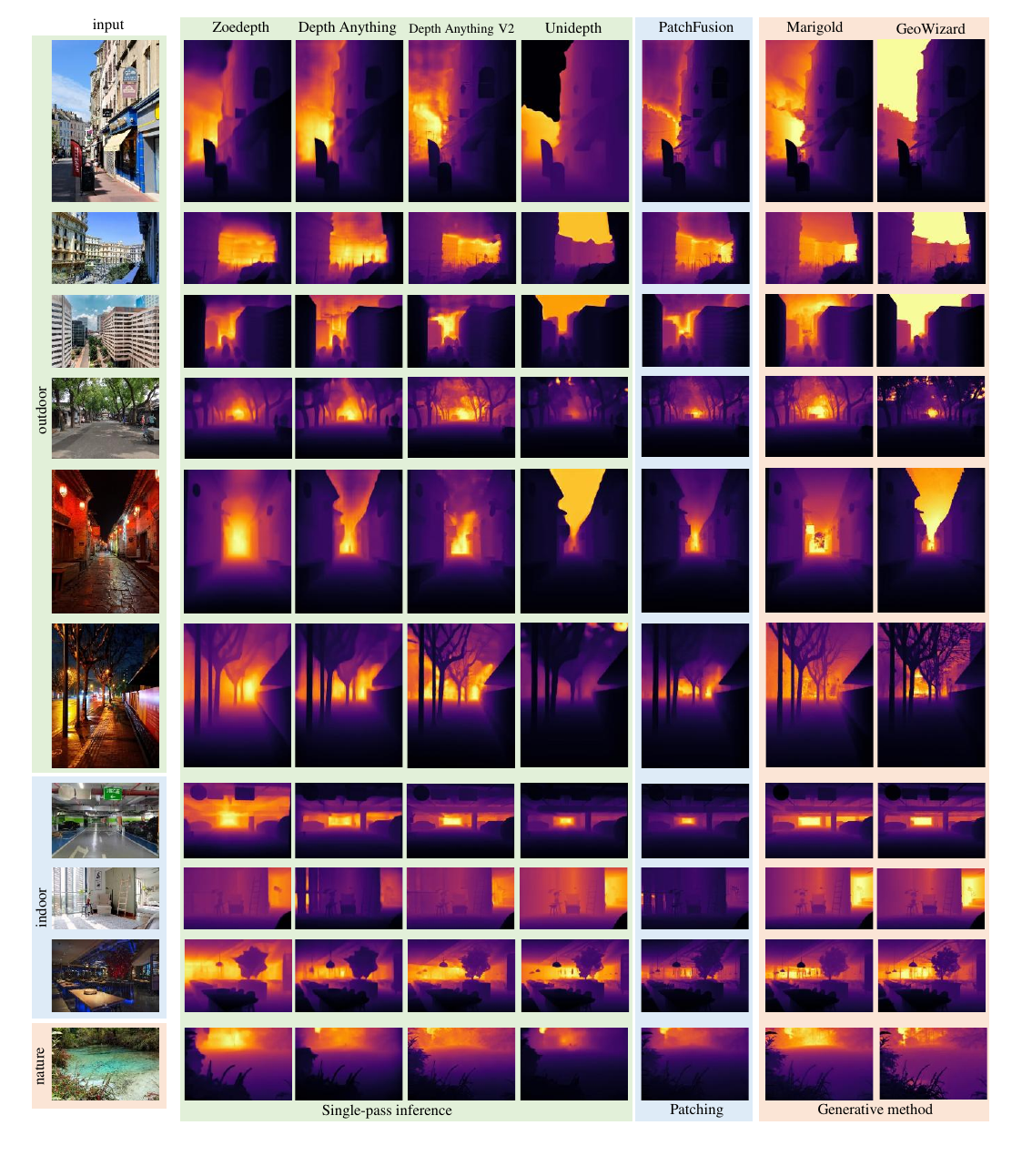}}
\caption{Performance comparison of MMDE models across diverse scenarios (indoor/outdoor, urban/natural, large/small scale, varying illumination). Colors indicate scene types and method categories. Generative methods predict relative depth, while others yield absolute depth~\citep{bochkovskii2024depth}.}
\end{figure*}

\begin{figure}
\centerline{\includegraphics[width=18pc]{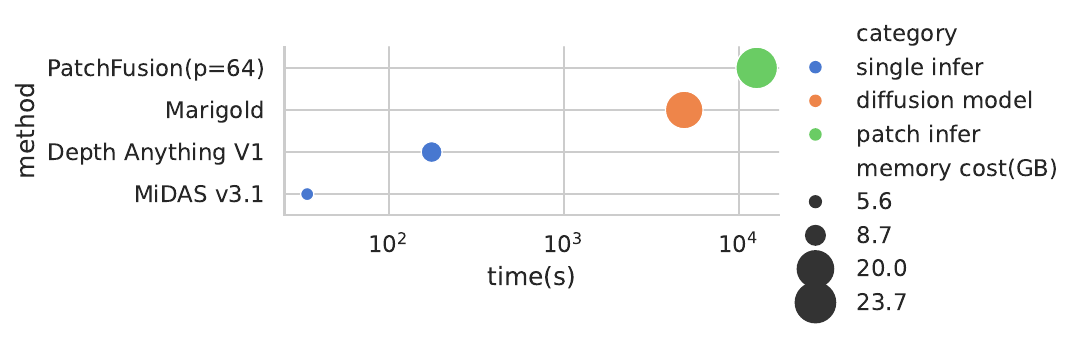}}
\caption{Inference time and memory usage for different model types are shown on a logarithmic scale in seconds.}
\end{figure}

Generative diffusion models offer an alternative by progressively refining depth predictions, capturing complex geometries with strong structural consistency~\citep{ke2024repurposing}. While effective in detail preservation and less reliant on labeled data, their multi-step denoising introduces computational overhead and variability. Most focus on relative depth, with limited exploration of monocular metric depth estimation (MMDE). Notably, DMD incorporates field-of-view conditioning and log-scale parameterization to achieve accurate zero-shot MMDE~\citep{saxena2023zero}, though inference inefficiency remains a bottleneck.

In summary, single-inference methods trade detail for speed, patch-based methods improve accuracy at high cost, and generative models offer superior fidelity but face scalability challenges.

\section{Datasets for MMDE}
A wide range of datasets, as shown in Table 3, supports monocular metric depth estimation (MMDE), with outdoor collections dominating due to their relevance to autonomous driving. Benchmarks such as KITTI, Waymo Open Dataset, and nuScenes provide rich multi-sensor data—including RGB, LiDAR, and radar—offering reliable metric depth for training and evaluation~\citep{geiger2013vision, caesar2020nuscenes, sun2020scalability}.

Indoor datasets, though fewer, are vital for robotics and AR applications. Notable examples include ScanNet, NYU Depth V2, and SUN RGB-D, which supply high-quality RGB-D imagery and accurate ground truth~\citep{dai2017scannet, silberman2012indoor, song2015sun}.

Synthetic datasets such as TartanAir, Hypersim, and vKITTI provide pixel-perfect, noise-free depth maps, complementing real-world datasets by enabling large-scale, diverse, and controlled training scenarios~\citep{wang2020tartanair, roberts2021hypersim, gaidon2016virtual}. These resources are increasingly leveraged to mitigate annotation errors and domain gaps in real-world data.

Dataset diversity also extends to sensor configurations, from RGB-only captures to multi-modal setups integrating LiDAR, GPS, and IMU. Crucially, most reviewed datasets (32 of 38) provide true metric depth, while a smaller subset offers only relative depth, which remains useful in scale-agnostic applications.

In summary, real-world driving datasets dominate current MMDE research, indoor datasets provide indispensable complementary supervision, and synthetic datasets enhance diversity and precision, together forming the backbone of benchmarking and training.

\section{Summary and Outlook}
MMDE has evolved from traditional architectures toward generative modeling and domain-generalizable frameworks. Tables 1 and 2 highlight art methods and performance benchmarks, illustrating progress in both architectural design and data utilization. These advances have broadened the applicability to 3D reconstruction, navigation, and interactive perception. Nonetheless, challenges remain, including fine-detail preservation, geometric consistency in complex scenes, and the trade-off between accuracy and efficiency~\citep{spencer2024third,yang2024depth1}.

Loss design continues to play a critical role, with edge-aware and gradient-based formulations improving structural fidelity, while generative supervision further enhances robustness~\citep{sun2025depth}. Data strategies combining real-world and synthetic datasets~\citep{wang2020tartanair,roberts2021hypersim} mitigate annotation scarcity and domain gaps, offering a scalable foundation for generalization.

Generative diffusion models, such as Marigold and GeoWizard, show strong potential in recovering high-frequency details and complex geometries, while DMD introduces field-of-view conditioning and log-scale parameterization for improved adaptability~\citep{bochkovskii2024depth}. Although diffusion methods remain computationally demanding, optimization of multi-step inference is advancing toward practical deployment.

A clear research trend is zero-shot generalization across unseen domains. Approaches like ZoeDepth and UniDepth~\citep{piccinelli2024unidepth} demonstrate promising transferability through architectural innovations and large-scale training. Looking forward, priorities include improving computational efficiency, enforcing geometric consistency in multi-view settings, and advancing domain adaptation.

MMDE is moving toward becoming a cornerstone of spatial perception, with ongoing innovations in loss functions, hybrid data pipelines, and generative modeling steadily pushing the field toward universal, accurate, and efficient depth estimation.

\captionsetup[table]{width=\textwidth,justification=raggedright}
\begingroup
\scriptsize
\begin{longtable}
{p{1.2cm}p{1cm}p{0.7cm}p{0.9cm}p{1.9cm}p{1.4cm}p{0.8cm}p{2.8cm}}
\caption{
Overview of depth estimation datasets commonly used in computer vision. Datasets providing metric depth (e.g., NYU-D, KITTI, ApolloScape) supply RGB–depth pairs for supervised MMDE, while relative depth datasets (e.g., DIW, Movies, WSVD) support ordinal depth prediction. Synthetic and hybrid datasets (e.g., BlendedMVS, TartanAir~\citep{wang2020tartanair}) augment real-world data and aid domain adaptation. Columns summarize dataset name, scene type (indoor/outdoor), driving relevance, synthetic/real origin, supported tasks, available modalities, supervision type, and key features, facilitating selection for applications ranging from autonomous driving to indoor scene understanding.}\\
\toprule
\textbf{Name} & \textbf{Indoor Outdoor} & \textbf{Driving Data} & \textbf{Synthetic Real} & \textbf{Tasks} & \textbf{Categories} & \textbf{Relative Metric} & \textbf{Description} \\
\midrule
\endfirsthead
 
Argoverse2 & Outdoor & Yes & Real &\tiny Trajectory Prediction, Object Detection, Depth Estimation, Semantic Segmentation, SLAM & RGB, LiDAR, GPS, IMU, 3D BBoxes, Labels & Metric&\tiny An autonomous driving dataset with 360-degree LiDAR and stereo data, focusing on long-term tracking and forecasting. \\
Waymo & Outdoor & Yes & Real &\tiny Object Detection, Depth Estimation, Semantic Segmentation, Trajectory Prediction, SLAM & RGB, LiDAR, GPS, IMU, 3D BBoxes, Labels & Metric&\tiny A large-scale autonomous driving dataset from urban and highway environments with multi-sensor data. \\
DrivingStereo & Outdoor & No & Real &\tiny Stereo Matching, Depth Estimation & High-Resolution Stereo Images, Depth Maps & Metric &\tiny High-resolution binocular images and ground truth depth maps for stereo vision tasks. \\
Cityscapes & Outdoor & No & Real &\tiny Semantic Segmentation, Instance Segmentation, Depth Estimation & RGB Images, Semantic Labels & Metric&\tiny A dataset of urban street scenes for semantic and instance segmentation. Depth data is typically derived. \\
BDD100K & Outdoor & Yes & Real &\tiny Object Detection, Semantic Segmentation, Driving Behavior Prediction, Depth Estimation & RGB, Semantic Labels, Videos, GPS, IMU & Relative&\tiny A large-scale driving dataset with diverse scenes and tasks, including object detection and behavioral prediction. \\
Mapillary Vistas & Outdoor & No & Real &\tiny Semantic Segmentation, Instance Segmentation, Depth Estimation & RGB Images, Semantic Labels, Depth Maps & Relative&\tiny A massive street-level imagery dataset with rich semantic annotations for segmentation and depth estimation. \\
A2D2 & Indoor Outdoor & Yes & Real &\tiny Semantic Segmentation, Object Detection, Depth Estimation, SLAM & RGB, Semantic Labels, LiDAR, IMU, GPS, 3D BBoxes & Metric&\tiny An autonomous driving dataset with multi-sensor data and detailed annotations, including both indoor and outdoor scenes. \\
ScanNet & Indoor & No & Real &\tiny 3D Reconstruction, Semantic Segmentation, Depth Estimation & RGB-D, Point Clouds, Semantic Labels & Metric&\tiny A large-scale indoor dataset with RGB-D images and 3D point clouds for 3D reconstruction and semantic understanding. \\
Taskonomy & Indoor Outdoor & No & Real &\tiny Multi-Task Learning, Depth Estimation, Semantic Segmentation & RGB, Depth Maps, Normals, Point Clouds & Metric&\tiny A dataset designed for multi-task learning, providing a wide range of visual tasks and their ground truth. \\
SUN-RGBD & Indoor & No & Real &\tiny 3D Reconstruction, Semantic Segmentation, Depth Estimation & RGB-D, Point Clouds, Semantic Labels & Metric&\tiny A dataset of indoor scenes providing RGB-D images for 3D reconstruction and depth estimation tasks. \\
Diode Indoor & Indoor & No & Real &\tiny Depth Estimation, 3D Reconstruction & RGB, LiDAR Depth Maps, Point Clouds & Metric&\tiny A high-precision dataset with LiDAR and RGB data for indoor depth estimation and 3D modeling. \\
IBims-1 & Indoor & No & Real &\tiny 3D Reconstruction, Depth Estimation & RGB-D, 3D Reconstruction & Metric&\tiny A dataset providing RGB-D images and 3D reconstructions of indoor building environments. \\
VOID & Indoor & No & Real &\tiny 3D Reconstruction, Depth Estimation, SLAM & RGB-D, Point Clouds & Metric&\tiny An RGB-D dataset of indoor scenes for 3D reconstruction, with a focus on occlusions. \\
HAMMER & Indoor Outdoor & No & Real &\tiny Depth Estimation, 3D Reconstruction, Semantic Segmentation, SLAM & RGB-D, Point Clouds, Semantic Labels & Metric&\tiny A high-precision dataset for depth estimation and 3D reconstruction, covering both indoor and outdoor scenes. \\
ETH-3D & Indoor Outdoor & No & Real &\tiny Multi-View Stereo, Depth Estimation, 3D Reconstruction & High-Res RGB, Depth Maps, Point Clouds & Metric&\tiny A benchmark for multi-view stereo and 3D reconstruction with high-resolution images from indoor and outdoor scenes. \\
nuScenes & Outdoor & Yes & Real &\tiny Object Detection, Trajectory Prediction, Depth Estimation, SLAM & RGB, LiDAR, Radar, GPS, IMU, 3D BBoxes, Labels & Metric&\tiny An autonomous driving dataset with a full sensor suite, including RGB, LiDAR, and radar data. \\
DDAD & Outdoor & Yes & Real &\tiny Depth Estimation, Object Detection, 3D Reconstruction, SLAM & RGB-D, Point Clouds & Metric&\tiny An autonomous driving dataset with rich sensor data and annotations, focused on dense depth and 3D reconstruction. \\
BlendedMVS & Indoor Outdoor & No & Synthetic Real &\tiny Multi-View Stereo, Depth Estimation, 3D Reconstruction & RGB, Depth Maps, 3D Reconstruction & Metric&\tiny A multi-view stereo dataset that blends real and synthetic data for depth estimation and 3D reconstruction. \\
DIML & Indoor & No & Real &\tiny Stereo Matching, Depth Estimation & RGB Images, Depth Maps & Metric&\tiny A multi-view dataset of indoor scenes designed for depth estimation and stereo matching. \\
HRWSI & Outdoor & No & Real &\tiny Stereo Matching, Depth Estimation & High-Resolution RGB, Depth Maps & Metric&\tiny A high-resolution dataset for stereo matching, providing RGB images and depth maps of outdoor scenes. \\
IRS & Indoor & No & Real &\tiny 3D Reconstruction, Depth Estimation, SLAM & RGB-D, Point Clouds & Metric&\tiny An indoor RGB-D dataset for depth estimation and 3D reconstruction tasks. \\
MegaDepth & Outdoor & No & Real &\tiny 3D Reconstruction, Depth Estimation & High-Res RGB, Depth Maps & Relative&\tiny A large-scale outdoor dataset for monocular depth estimation, providing high-resolution images and relative depth maps. \\
TartanAir & Indoor Outdoor & No & Synthetic &\tiny SLAM, Depth Estimation, 3D Reconstruction, Semantic Segmentation & RGB, Depth Maps, Point Clouds, Labels & Metric&\tiny A synthetic dataset for visual SLAM and depth estimation, offering diverse indoor and outdoor scenes. \\
Hypersim & Indoor Outdoor & No & Synthetic &\tiny 3D Reconstruction, Scene Understanding, Semantic Segmentation & RGB, Depth Maps, Point Clouds, Labels & Metric&\tiny A high-precision synthetic dataset of indoor and outdoor scenes for 3D reconstruction and scene understanding. \\
vKITTI & Outdoor & Yes & Synthetic &\tiny Object Detection, Semantic Segmentation, Depth Estimation, SLAM & Synthetic RGB, Labels, Depth Maps, 3D BBoxes & Metric&\tiny A virtual driving dataset offering synthetic scenes with full ground truth for tasks like object detection and depth estimation. \\
KITTI & Outdoor & Yes & Real &\tiny Object Detection, Stereo Matching, Depth Estimation, SLAM & RGB, LiDAR, Depth Maps, Semantic Labels & Metric&\tiny A foundational autonomous driving dataset with multi-sensor data for a wide range of computer vision tasks. \\
NYU-D & Indoor & No & Real &\tiny Semantic Segmentation, Depth Estimation, 3D Reconstruction & RGB-D, Semantic Labels, Point Clouds & Metric&\tiny A large-scale indoor RGB-D dataset widely used for depth estimation and semantic segmentation. \\
Sintel & Outdoor & No & Synthetic &\tiny Optical Flow, Depth Estimation, Semantic Segmentation & RGB, Depth Maps, Labels, Optical Flow & Metric&\tiny A synthetic movie-like dataset for optical flow, depth estimation, and other visual tasks. \\
ReDWeb & Outdoor & No & Real &\tiny Monocular Depth Estimation & RGB Images & Relative&\tiny A dataset for monocular depth estimation, using web videos to supervise relative depth prediction. \\
Movies & Indoor Outdoor & No & Synthetic Real &\tiny Monocular Depth Estimation, Zero-Shot Learning & RGB Images & Relative&\tiny A blended dataset used for zero-shot monocular depth estimation from various sources. \\
ApolloScape & Outdoor & Yes & Real &\tiny Object Detection, Semantic Segmentation, Depth Estimation & RGB, LiDAR & Metric&\tiny A large-scale autonomous driving dataset with extensive annotations for segmentation and depth estimation. \\
WSVD & Indoor Outdoor & No & Real &\tiny Monocular Depth Estimation & RGB Images & Metric&\tiny A web video-supervised dataset for dynamic scene depth prediction, emphasizing moving objects. \\
DIW & Outdoor & No & Real &\tiny Monocular Depth Estimation & RGB Images & Relative&\tiny A dataset providing relative depth supervision for a wide range of outdoor scenes. \\
ETH3D & Indoor Outdoor & No & Real &\tiny Multi-View Stereo, Depth Estimation & High-Res RGB, Videos & Metric&\tiny A benchmark for multi-view stereo and depth estimation with high-resolution images. \\
TUM & Indoor & No & Real &\tiny RGB-D SLAM & RGB-D & Metric &\tiny A dataset for evaluating RGB-D SLAM algorithms and depth estimation in indoor environments. \\
3D Ken Burns & Indoor & No & Synthetic &\tiny Depth Estimation, Image Animation & Static Images, Depth Maps & Metric&\tiny A dataset for generating animated videos with a "Ken Burns" effect using depth information. \\
Objaverse & Indoor Outdoor & No & Synthetic Real&\tiny 3D Object Detection, Classification & 3D Models & Relative&\tiny A vast dataset of 3D objects for recognition, understanding, and generation tasks. \\
OmniObject3D & Indoor & No & Synthetic &\tiny 3D Object Detection, Recognition, Reconstruction & 3D Models, RGB-D & Metric&\tiny A synthetic dataset with multi-view images and corresponding depth information for 3D object tasks. \\
 
\bottomrule
\end{longtable}
\endgroup

\bibliography{nips2022}
\bibliographystyle{icml2022}


\end{document}